%% file: main.tex
\documentclass[10pt,twocolumn,letterpaper]{article}

\usepackage[pagenumbers]{cvpr} 

\input{sec/preamble}

\usepackage{epigraph} 

\definecolor{cvprblue}{rgb}{0.21,0.49,0.74}
\usepackage[pagebackref,breaklinks,colorlinks,citecolor=cvprblue]{hyperref}

\def\paperID{466} 
\def\confName{CVPR}
\def\confYear{2024}

\title{AutoAD III: The Prequel -- Back to the Pixels}

\author{
Tengda Han$^{1}$ \quad Max Bain$^{1}$ \quad
Arsha Nagrani$^{1\dagger}$ \quad G\"ul Varol$^{1,2}$ \quad Weidi Xie$^{1,3}$ \quad Andrew Zisserman$^1$\\
{\small$^1$Visual Geometry Group, University of Oxford} \quad
{\small$^2$LIGM, \'Ecole des Ponts ParisTech}
\quad{\small$^3$CMIC, Shanghai Jiao Tong University} \\
{\small\url{https://www.robots.ox.ac.uk/vgg/research/autoad/}}\vspace{-5mm}
}

\begin{document}
\maketitle
\input{sec/0_abstract}    
\input{sec/1_intro}
\input{sec/2_related_works}

\input{sec/3_method}

\input{sec/4_exp}

\input{sec/9_conclusion}

\clearpage
{
    \small
    \bibliographystyle{ieeenat_fullname}
    \bibliography{bib/shortstrings,bib/vgg_local,bib/vgg_other}
}

\input{sec/X_appendix}

\end{document}

%% file: sec/preamble.tex
\usepackage[dvipsnames]{xcolor}
\newcommand{\blue}[1]{{\color{black}#1}}

\newcommand{\app}[0]{{Appendix}}

\usepackage{pifont}
\newcommand{\cmark}{\ding{51}}%
\newcommand{\xmark}{\ding{55}}%

\DeclareMathOperator*{\argmin}{arg\,min} 
\DeclareMathOperator*{\argmax}{arg\,max} 
\usepackage{multirow}

\usepackage[subtle]{savetrees} 

\usepackage[ruled]{algorithm2e}
\usepackage{listings}
\DeclareFixedFont{\ttb}{T1}{txtt}{bx}{n}{6} 
\DeclareFixedFont{\ttm}{T1}{txtt}{m}{n}{6}  

\definecolor{deepblue}{rgb}{0,0,0.5}
\definecolor{deepred}{rgb}{0.6,0,0}
\definecolor{deepgreen}{rgb}{0,0.5,0}

\definecolor{codegreen}{rgb}{0,0.6,0}
\definecolor{codegray}{rgb}{0.5,0.5,0.5}
\definecolor{codepurple}{rgb}{0.58,0,0.82}
\definecolor{backcolour}{rgb}{0.95,0.95,0.92}

\definecolor{codeblue}{rgb}{0.25,0.5,0.5}
\usepackage{float}

%% file: sec/0_abstract.tex
\begin{abstract}
Generating Audio Description (AD) for movies is a challenging task that requires fine-grained visual understanding and an awareness of the characters and their names. Currently, visual language models for AD generation are limited by a lack of suitable training data, and also their evaluation is hampered by using performance measures not specialized to the AD domain. In this paper, we make three contributions: (i) We propose two approaches for constructing AD datasets with aligned video data, and build training and evaluation datasets using these. These datasets will be publicly released; (ii) We develop a Q-former-based architecture which ingests raw video and generates AD, using frozen pre-trained visual encoders and large language models; and (iii) We provide new evaluation metrics to benchmark AD quality that are well matched to human performance.
Taken together, we improve the state of the art on AD generation.
\end{abstract}

%% file: sec/1_intro.tex
\section{Introduction}
\label{sec:intro}

\renewcommand{\epigraphflush}{flushleft}
\renewcommand{\epigraphsize}{\small}
\setlength{\epigraphwidth}{0.45\textwidth}
{\scriptsize
\epigraph{\textit{Cinema is a matter of what's in the frame and what's out.}\\
\hspace{150pt}\textit{Martin Scorsese}}
{}
}\vspace{-3mm}

Audio description (AD) is an accessibility tool for the blind and visually impaired that describes visual content
which is essential for following video programs\footnote{\small\url{https://www.3playmedia.com/learn/popular-topics/audio-description/}}.
Automatically generating AD text is a challenging task as the information must be accurate, character-aware, story-aware, complementary to the soundtrack, and distilled into the gaps between speech. For TV broadcasters in the US and UK, providing AD for a certain percentage of video content has become a legal requirement.

With the current power of visual-to-text generative models, generating AD automatically is now becoming possible \cite{wang2021toward}, and there has been a recent flurry of interest in this goal~\cite{Han23,Han23a},
kick-started by the availability of films and AD provided in the MAD dataset~\cite{soldan2022mad}. Key innovations have included partial training of AD generative models using available large-scale datasets~\cite{Han23}, and the introduction of a {\em character bank} to provide hints (as prompts) for the language model for the crucial objective of naming the characters in the generated text descriptions~\cite{Han23a}. 
However, MAD only provides frame-level CLIP features (and only at 5 Hz) and this has limited the ability of generative models to provide fine-grained spatial details. Recent Visual Language Models (VLMs)~\cite{alayrac2022flamingo,li2022blip,li2023blip2,Li2023Lavander,yu2022coca,MERLOT} have accessed the spatial feature map of the image (or video) in order to obtain fuller descriptions or answer more detailed questions.

\begin{figure} 
    \centering
    \includegraphics[width=1\linewidth]{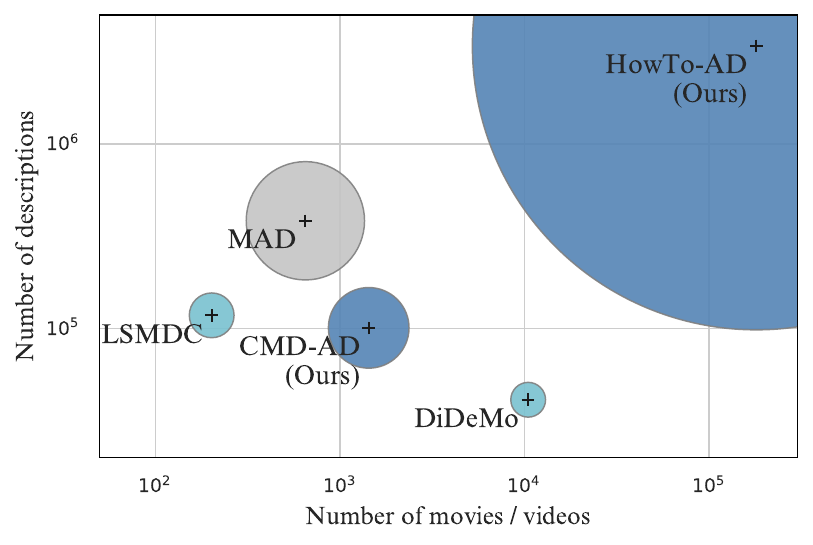}
    \vspace{-0.8cm}
    \caption{We propose two new movie Audio Description (AD) datasets \emph{with pixels} -- \textbf{CMD-AD} and \textbf{HowTo-AD} by temporally aligning or textually transforming existing pixel video datasets. The marker size is proportional to the total video durations and {\color{gray} grey color} indicates datasets with features instead of raw pixels.
    } 
    \vspace{-0.5cm}
    \label{fig:teaser}
\end{figure}

We make the following contributions: 
First, we provide {\em two new datasets} that can be used to train an AD generation model end-to-end. The datasets go beyond MAD~\cite{soldan2022mad} in that they provide video as input, rather than only a CLIP frame feature, i.e., they go {\em back to the pixels}. The first dataset, {\em CMD-AD}, is constructed from two publicly available sources -- the AD descriptions for films available from AudioVault\footnote{\url{https://audiovault.net}}
and the movie clips available from CMD~\cite{Bain20}. The challenge in this case is how to determine the temporal alignment of these two sources given that one has only audio with AD, and the other (CMD) is non-contiguous with timings unknown with respect to the original movies. The second dataset, {\em HowTo-AD}, is constructed from the large-scale HowTo100M video dataset~\cite{miech2019howto100m}
that originally consists of YouTube videos with narrated instructions. Inspired by the use of Language Models (LMs) to rephrase the instructions as video captions in HowtoCaption~\cite{shvetsova2023howtocaption},
we use LMs to repurpose HowTo100M as an AD dataset containing videos with an associated character bank, and text descriptions of the visual content that also names the person performing the actions. While this dataset is not a ground-truth AD dataset, we show that the pseudo ground-truth annotations are a valuable source of training data for AD. The statistics of these two new datasets are given in Table~\ref{tab:ad_datasetsl}, and visually illustrated in Figure~\ref{fig:teaser}.

Our second contribution is a new architecture for AD generation that directly inputs a video clip and character bank proposals, and outputs a character-aware description. The model is based on the Q-former architecture of BLIP-2~\cite{li2023blip2} that bridges the visual space with the language space,
then generates textual outputs with a large language model~\cite{zhu2023minigpt,yu2023videoblip,damonlpsg2023videollama}.
Our architecture is different from BLIP-2~\cite{li2023blip2} in that (i) it takes multi-frame movie clips as visual inputs, 
and (ii) it incorporates character bank information both from the face exemplars and the character names.

\begin{table} 
    \centering
    \footnotesize
    \setlength{\tabcolsep}{2pt}
    \begin{tabular}{ccrrr}
    \toprule
        Dataset & with pixels & \# movies & \# AD & total duration\\ \midrule
        \textcolor{gray}{MovieNet~\cite{huang2020movienet_short}}  & \textcolor{gray}{\cmark}    & \textcolor{gray}{1100} & \textcolor{gray}{--} & \textcolor{gray}{--} \\
        LSMDC~\cite{rohrbach2015lsmdc}   & \cmark  & 202  & 118k & 147h \\
        MAD~\cite{soldan2022mad}         & \xmark  & 650  & 384k & 1027h \\
        CMD~\cite{Bain20}                & \cmark  & 3605 & --  & 1270h \\
        CMD~\cite{Bain20} $\cap$ AudioVault-8K~\cite{Han23} & \cmark & 1803 & --  & 647h \\ \midrule
        \textbf{CMD-AD (ours)} & \cmark  & 1432 & 101k & 477h \\
        \textbf{HowTo-AD (ours)} & \cmark  & 180,034* & 3.4M & 23652h  \\
    \bottomrule
    \end{tabular}
    \vspace{-0.2cm}
    \caption{\textbf{Statistics of Movie AD datasets.}
    Only a small number of movie datasets with AD are available, 
    and they have different limitations:
    MovieNet only provides keyframes, LSMDC is short in duration, MAD only provides frame-level visual features, and CMD does not have corresponding ADs. We propose two new datasets for AD generation task: CMD-AD and HowTo-AD. *: strictly they are long videos rather than movies.
    }
    \vspace{-0.5cm}
    \label{tab:ad_datasetsl}
\end{table}

Our third contribution is on {\em evaluation}. Previous methods use a small test set of only 10 movies for evaluation. We introduce an evaluation dataset of 100 movies, based on our aligned movie clips with AD from AudioVault. This has far more {\em diversity} than the previous test sets used, covering, e.g.\ science fiction, westerns, action, horror, cartoon, and romance.
As well as introducing a new test set, we also adopt two new evaluation methods. For AD, the gold standard evaluation is to compare the generated AD with that provided by humans. For model development, however, an automatic scalable evaluation is required. Previous works have used captioning metrics such as CIDEr~\cite{vedantam2015cider} but these have severe limitations since they essentially measure n-gram accuracy, and the same semantic AD can be presented in multiple equivalent ways. To deal with this problem, \cite{Han23a} introduced a retrieval-based assessment, evaluating how often we can pick out the correct AD out of multiple neighboring ADs by comparing them to the generated AD using BertScore~\cite{zhang2019bertscore} semantic text similarity.
In this work we adopt two new measures. The first called CRITIC, addresses one essential element of AD that distinguishes it from standard video captioning -- that it must name the characters involved. The second measure follows the recent trend in using LLMs to assess the veracity of captioning~\cite{zheng2023judging,chiang2023can,chan2023clair,maaz2023video,song2023moviechat}
As an exemplar of the usefulness of these new measures we also use them to assess inter-rater consistency where the same film has AD provided by several human annotators.
On these and traditional metrics, we show that our new architecture trained on raw pixels directly achieves impressive results for the task of Movie AD, outperforming previous works on both the standard MAD~\cite{soldan2022mad} eval set, and our new proposed test set. 

%% file: sec/2_related_works.tex
\vspace{-0.5mm}
\section{Related Work}
\label{sec:related_works}

\noindent\textbf{Dense video captioning.}
With the availability of large-scale data, the field has made significant
progress in captioning images~\cite{yu2022coca,mokady2021clipcap,li2022blip,li2023blip2},
and trimmed short video segments~\cite{sharma2018cc,lin2022swinbert,luo2020univilm,seo2022end}.
Movie AD generation is more related to the task of \textit{dense} video captioning,
where the goal is to concurrently address temporal localization and to describe
each identified interval in an untrimmed video~\cite{krishna2017dense}.
The approach to dense captioning has been explored either in two stages
\cite{krishna2017dense,iashin2020better,iashin2020multi,wang2018bidirectional,wang2020event} or a single stage~\cite{chadha2020iperceive,chen2021towards,deng2021sketch,li2018jointly,mun2019streamlined,rahman2019watch,shen2017weakly, shi2019dense,wang2018bidirectional,wang2021end,zhou2018end,yang2023vid2seq},
depending on whether localization and captioning are jointly addressed.
Standard evaluation benchmarks for dense captioning
consist of web videos such as
YouCook2~\cite{youcook2}, ActivityNet Captions~\cite{krishna2017dense}, and ViTT~\cite{huang2020multimodal}.
Recently, Vid2Seq~\cite{yang2023vid2seq} repurposed the narrated web videos YT-Temporal-1B~\cite{zellers2022merlotreserve},
using transcribed speech as the supervision source. In a similar spirit,
HowToCaption~\cite{shvetsova2023howtocaption} was collected by transforming the narrations of HowTo100M~\cite{miech2019howto100m} into caption-like descriptions using LLMs, and LaVila~\cite{zhao2023lavila}
captioned long videos to enable large-scale video-text pretraining, also leveraging LLMs.
The distinction between AD generation and dense captioning lies in
the former's focus on character names, story relevance, and avoidance of
interference with important audio content (\eg character speech).

\noindent\textbf{Movie understanding datasets.} For movie understanding, current datasets facilitate a range of computer vision tasks including metadata classification~\cite{pardo2022moviecuts}, VQA~\cite{tapaswi2016movieqa}, and visual character grounding~\cite{rohrbach2017movie}. These tasks often rely on auxiliary data such as movie plots~\cite{xiong2019graph}, book adaptations~\cite{tapaswi2015book2movie}, or AD~\cite{soldan2022mad} for dense annotations. However, due to copyright constraints, many datasets are limited to offering visual features (MAD~\cite{soldan2022mad}) or sparse keyframes (MovieNet~\cite{huang2020movienet_short}). CMD~\cite{Bain20} circumvents this by providing urls to licensed YouTube clips. AutoAD~\cite{Han23} improves on automatic AD collection, and provides large-scale audio AD data.

\noindent\textbf{Improvements in VLMs for images and videos.}
The recent success of LLMs~\cite{touvron2023llama2,chowdhery2022palm,zhang2022opt,touvron2023llama} and vision encoders~\cite{dosovitskiy2020image,clip2021} has led to an explosion of multimodal (vision and language) models that can jointly understand both vision and text data.
These methods largely work by mapping \textit{frozen} image encoders (e.g.\ CLIP~\cite{clip2021}, EVA-CLIP~\cite{fang2023eva}) to the textual embedding space of \textit{frozen} LLMs~\cite{touvron2023llama2,zhang2022opt,touvron2023llama}, for example Flamingo~\cite{alayrac2022flamingo}, which does so via a Perceiver resampler~\cite{jaegle2021perceiver}, or BLIP2~\cite{li2023blip2}, which uses a Q-former to achieve a similar mapping. Video-LLama~\cite{damonlpsg2023videollama} extends this idea to the audiovisual domain, by using the multimodal ImageBind ~\cite{girdhar2023imagebind} encoder in conjunction with video and audio Q-formers. 
While MV-GPT~\cite{seo2022end} finetunes a native video backbone~\cite{arnab2021vivit} for the task of video captioning, most works adapt image encoders to the costly video domain via temporally sampling a few frames with large strides~\cite{wang2022git, chen2023pali}, or by 
representing each frame by
a single token~\cite{yang2023vid2seq, zhou2018end, wang2021end}. Given the impressive generalisation capabilities of these works made of up strong frozen components~\cite{yu2023self}, we also adopt a similar approach, leveraging Video-LLama~\cite{damonlpsg2023videollama} and BLIP-2~\cite{li2023blip2} models as our backbone, with the key addition that we also integrate character information. 
More recent works such as MiniGPT-4~\cite{zhu2023minigpt},  MovieChat~\cite{song2023moviechat} and VideoBLIP~\cite{yu2023videoblip} use stronger instruction-tuned LLMs, enabling further zero-shot capabilities.

\noindent\textbf{Captioning evaluation.}
Human evaluation is the gold standard for judging caption quality, however it requires multiple annotators for consistency, is expensive and exceptionally slow. Existing automatic metrics, such as BLEU, ROUGE and CIDEr~\cite{vedantam2015cider}, all primarily measure n-gram overlap (however have different weighting schemes between n-grams, and across precision/recall), and do not capture the inherent subjectivity of the task, where different phrasing is often equally valid. Other metrics include SPICE~\cite{anderson2016spice} (adds action and
object relationships), while model-based metrics using earlier language models or image-language models include BERT-Score~\cite{zhang2019bertscore}, BERT-Score++~\cite{yi2020improving} (fine-tunes
BERT for image captioning), LEIC~\cite{cui2018learning} 
and NUBIA~\cite{kane2020nubia} (custom trained models for image caption evaluation), TIGEr~\cite{jiang2019tiger}, CLIPScore~\cite{hessel2021clipscore}, \blue{and EMScore~\cite{shi2022emscore}}. Given the explosion of LLMs, however, recent works explore the use of state-of-the-art LLMs, such as GPT-4, as a surrogate for humans. Because these models are often trained with RLHF, they already exhibit strong human alignment~\cite{bubeck2023sparks}, and can be used to assess text quality well (LLM-as-a-judge).~\cite{zheng2023judging, chiang2023can} show that using strong LLMs as judges (such as GPT-4) aligns highly with human preferences on a range of standard language-based tasks, such as conversational instruction following. CLAIR~\cite{chan2023clair} extends this idea to image captioning, showing similar strong correlations to human preferences on visual-language datasets such as MS-COCO and Flickr8K, while VideoChatGPT~\cite{maaz2023video} and MovieChat~\cite{song2023moviechat} use LLM-assisted evaluation for video tasks such as videoQA as well.

%% file: sec/3_method.tex
\section{New Datasets for Pixels to AD}
\label{sec:method}
In this section, we describe our two new datasets that contain raw video pixels mapped to AD annotation: CMD-AD (Sec.~\ref{subsec:data:cmd}) which is based on CMD~\cite{Bain20}, and HowTo-AD (Sec.~\ref{subsec:data:howtoad}) based on HowTo100M~\cite{miech2019howto100m}.

\subsection{CMD-AD -- Pixels from Aligned CMD}
\label{subsec:data:cmd}

The AudioVault website provides human annotated Audio Descriptions in the form of audio files with the spoken AD added to the original movie soundtrack (no video).
The CMD dataset~\cite{Bain20} consists of short (about 2 minutes long) non-contiguous movie clips in the form of video files on YouTube (around 10 clips per movie).
Although there are about 2000 movies overlapping between these two data sources,
temporally aligning the AD with the movie clips from CMD is a non-trivial task due to several challenges:
First, the movie soundtracks from AudioVault audio files have been modified and re-encoded to add the AD, therefore the audio signals from AudioVault files and CMD movie clips are not identical;
second, AudioVault audio files cover the full movie duration (e.g.\ around 90 minutes), whilst a CMD clip covers only 2 minutes, and performing precise alignment over the extent of the movie has the potential for many erroneous matches across the search space;
third, the same movie published in different locations might have been recorded at different speeds (e.g.\ NTSC 29.97 fps vs.\ PAL 25 fps\footnote{\url{https://en.wikipedia.org/wiki/576i\#PAL_speed-up}}), introducing another unknown into the alignment.

We propose a two-stage alignment pipeline to overcome these challenges and get precise temporal alignment between hour-long AudioVault audio files and non-contiguous short CMD movie clips from the same movie. To achieve this, we use two quasi-independent modalities: (i) the transcribed spoken text from the characters (not the AD), and (ii) the raw audio signal containing both non-speech sounds (music, sound effects) and the speech.

\begin{table} 
    \centering
    \footnotesize
    \setlength{\tabcolsep}{3pt}
    \begin{tabular}{crr}
    \toprule
        Split             & \# movies & \# AD \\ \midrule
        CMD-AD-Train & 1332 & 93,952 \\
        CMD-AD-Eval  & {100} & 7,316 \\
        total             & 1432 & 101,268 \\
    \bottomrule
    \end{tabular}
    \vspace{-0.2cm}
    \caption{\textbf{Statistics of the CMD-AD dataset.}}
    \vspace{-0.6cm}
    \label{tab:aligned_cmd}
\end{table}

\vspace{0.2cm}
\noindent\textbf{Stage1: Text-text alignment.} The aim in this stage is to first roughly localize the CMD movie clip with the AudioVault audio to reduce the search space.
In detail, we use WhisperX~\cite{Bain23} with the diarization module to separate the AD narration from the character speech, and obtain movie `subtitles' with timestamps for both AudioVault audio and CMD movie clips.
These are denoted as $\mathcal{S}_{\text{AV}}=\{(s_1,t_1),...,(s_m,t_m)\}$ and 
$\mathcal{S}_{\text{CMD}}=\{(s'_1,t'_1),...,(s'_n,t'_n)\}$,
where each $s_i$ denotes subtitle strings and $t_i$ denotes the temporal extent of this subtitle. Note that $n\ll m$ because CMD movie clips are much shorter than the entire movie, also the subtitles from the two sources are different because of arbitrary sentence partitioning by WhisperX or possible diarization errors.
To localize the CMD clip on the AudioVault movie time axis, 
we compute a simple word-error-rate (WER) using a sliding window approach as follows: we combine the CMD subtitles into a paragraph $\mathcal{P}_{\text{CMD}}=[s'_1;...;s'_n]$, then compute WER with AudioVault subtitles within a chunk size of $n$.
Formally, let $\mathcal{P}_{\text{AV}}^{(i)}=[s_i;...;s_{i+n}]$ denote the AudioVault subtitle paragraph consisting of $n$ continuous subtitle entries starting from $i$-th subtitles. For a particular CMD clip, the objective of text-text alignment is
\vspace{-1mm}
\begin{equation}
T_{\text{tt-align}} = \argmin_{t_i} \Bigl\{ \text{WER}(\mathcal{P}_{\text{CMD}},\mathcal{P}_{\text{AV}}^{(i)}) \Bigl\}.
\vspace{-1.5mm}
\end{equation}
The text-text alignment is not accurate when the CMD movie clip does not have many dialogues,~\eg~in action movies. 
In practice, we find it gives reliable rough time points for more than 90\% of CMD clips by randomly checking 10+ movies manually.

\begin{figure} 
    \centering
    \includegraphics[width=0.48\textwidth]{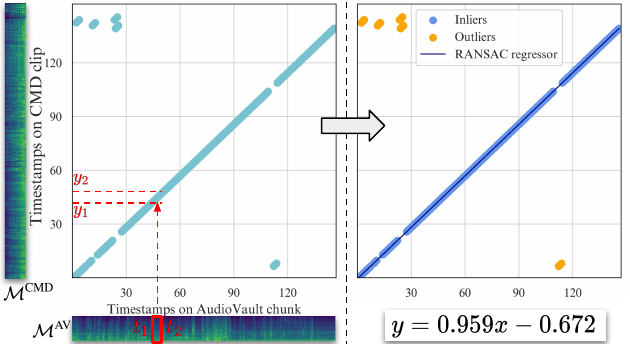}
    \vspace{-0.4cm}
    \caption{\textbf{Audio-audio alignment between two sources.}
    \textbf{(left)}:~For each small audio segment on AudioVault, we find the best-matching audio segment on CMD clip, and plot two timestamps as scatters;
    \textbf{(right)}:~Fitting a straight line with RANSAC we can get the precise mapping function between two sources.
    The slope of the fitted line $0.959<1$ indicates this CMD clip plays slightly faster than the corresponding AudioVault chunk.}
    \vspace{-0.3cm}
    \label{fig:method:ransac}
\end{figure}

\vspace{0.2cm}
\noindent\textbf{Stage2: Audio-audio alignment.} Given the rough alignment (which may be noisy) provided by Stage 1, this stage aims to verify the match, and obtain a \textit{precise} temporal alignment by comparing audio signals from the two sources.
The objective is to get a precise linear mapping for each CMD movie clip:
\vspace{-1mm}
\begin{equation}
    f: \{T_{\text{AV}} \rightarrow T_{\text{CMD}}\} = W\cdot t_{\text{AV}} + B,
\label{eq:method:ransac}
\vspace{-1mm}
\end{equation}
where $W$ is the speed rate between the AudioVault and CMD movie sources which might not be $1.0$ due to different movie fps,
and $B$ is the time shift. The key idea here is that even though the individual CMD clips are matched locally, the parameters $W$ and $B$ can be assumed to be global (i.e.\ constant) across the movie. Hence, matches can be verified as they will lie on a line specified by  $W$ and $B$, and this line can be obtained by a standard robust fitting method. Here we use RANSAC~\cite{Fischler81}.

To obtain precise audio alignment, 
we perform alignment on low-level audio representation mel-spectrogram.
First, we compute mel-spectrogram for both AudioVault audio and the CMD movie clip, denoted as $\mathcal{M}^{\text{AV}}$ and $\mathcal{M}^{\text{CMD}}$.
We only take a short AudioVault audio chunk based on the previous text-text alignment result. 
Second, we mask out mel-spectrogram regions of Audio Descriptions based on the timestamps obtained from WhisperX, as the AD signal only exists in AudioVault and not in the 
CMD movie clip.
Next, we perform sliding window matching with a window size $w=1.6s$. For each 1.6-second audio chunk on AudioVault starting from $t_1$ to $t_2$, 
we find the corresponding timestamps on CMD audio which has a maximum correlation:
\vspace{-1mm}
\begin{equation}
    y_1, y_2 = \argmax_{t_i, t_i+w} \Bigl\{
    \text{cor}\Big(
        \mathcal{M}^{\text{AV}}_{[t_1,t_2]}, \mathcal{M}^{\text{CMD}}_{[t_i,t_i+w]}
        \Big)
    \Bigl\}.
\vspace{-1mm}
\end{equation}
These matches can be thought of as points on a scatter plot from $(t_1,y_1)$ to $(t_2,y_2)$ for a series of small windows from AudioVault, as shown in Figure~\ref{fig:method:ransac}.
Finally, we use a RANSAC algorithm to fit a line through these match points over all clips to obtain the mapping in Equation~(\ref{eq:method:ransac}).
Based on the ratio between common movie fps, we filter RANSAC output by $0.8<W'<1.25$ and empirically choose mean-square-error $\text{MSE} < 100$. We find these two conditions give very decisive boundaries for confident RANSAC output. For instance, the successful RANSAC fitting at Figure~\ref{fig:method:ransac} has an MSE of $0.68$, whereas failed fittings typically have an \blue{MSE $>500$}.

\vspace{0.2cm}
\noindent\textbf{Summary.}
With this two-stage method,
we obtain accurate temporal alignment between AudioVault audio and CMD movie clips, therefore we can map the AudioVault AD annotations onto the CMD time axis to get video-text annotations. This gives us the dataset CMD-AD (statistics are provided in Table~\ref{tab:aligned_cmd}), consisting of 101k AD segments spanning 1,432 movies.
Note that the total number of overlapping movies between the two datasets is 1,803,
which means an 80\% success rate of precise alignment.
A higher success rate can be achieved by using a larger search window or an iterative alignment pipeline, which we leave as future work.

We use 1332 movies for training and 100 movies for evaluation,
naming the splits CMD-AD-Train and CMD-AD-Eval sets, respectively.

\begin{figure*}
    \centering
    \includegraphics[width=0.9\textwidth]{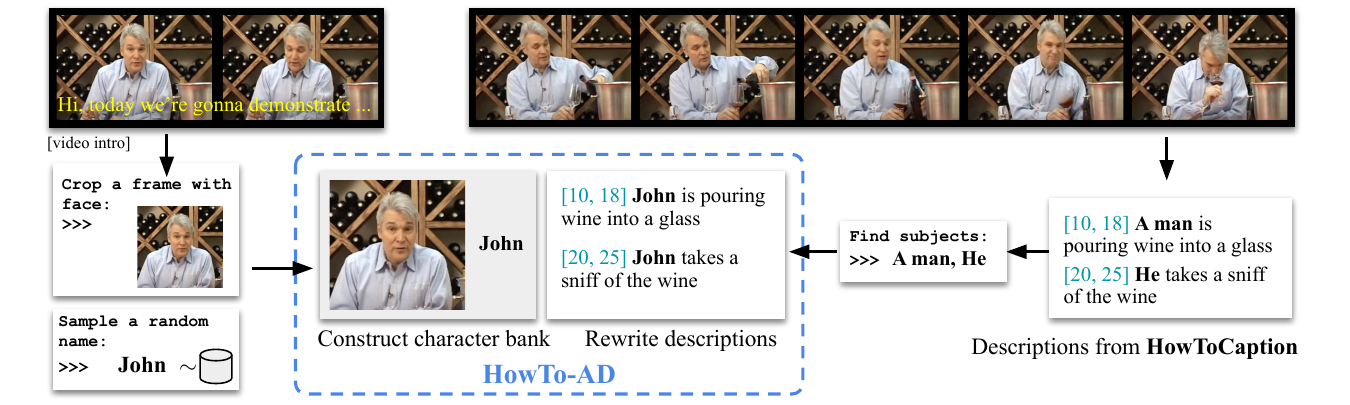}
    \vspace{-0.3cm}
    \caption{
    \textbf{HowTo-AD dataset}. We convert the LLM rewritten video descriptions (from HowToCaption) to fit movie audio descriptions by (i) uniformly replacing the subjects in descriptions with a randomly sampled name,~\ie \textbf{John}, and (2) constructing a character bank by providing a frame with the instructor and the randomly sampled name.
    The video sample is from~{\small\url{https://youtu.be/aRbQb19v2JI}}.}
    \vspace{-0.5cm}
    \label{fig:howtoad}
\end{figure*}

\subsection{HowTo-AD -- Pixels from HowTo100M}
\label{subsec:data:howtoad}

Our second dataset is based on the large-scale instructional video dataset HowTo100M~\cite{miech2019howto100m}, that contains over 1.2M videos with ASR subtitles from YouTube. At first glance, the ASR transcripts of these videos may look drastically different from that of AD in movies, since the spoken words are primarily aimed to instruct the viewer on how to carry out various daily tasks.

However, we can 
 {\em transform} the instruction ASR into pseudo-AD in two steps.
The first step is to adopt the captions generated from HowToCaption~\cite{shvetsova2023howtocaption},
where the ASR transcripts have been transformed into concise and {\em descriptive captions} with large language models~(LLMs). To improve caption temporal alignment with the corresponding video timestamps, 
the authors employ an off-the-shelf Temporal Alignment Network~\cite{Han22a}, 
while also discarding non-alignable subtitles (such as ``Hello, welcome to my channel!").
The second step addresses the key difference between descriptive captions and audio descriptions, that is, character names do not appear in the captions. For this transformation, we detect the subjects of description sentences and uniformly replace them with a randomly chosen character name,~\eg~transforming `\emph{a man} is pouring wine' into `\emph{John} is pouring wine'.
This completes the transformation from HowTo100M captions to the HowTo-AD.

Additionally, to mimic having a {\em character bank} as external knowledge as in~\cite{Han23a,yang2023dawn,Lin2023mmvid}, we also provide each instructional video with a pseudo-character bank that includes: the chosen character name and the character portrait face extracted from the instructional video, and a few face exemplars sampled from other videos to mimic  off-screen characters. \
An overview of the pipeline with an example is shown in Figure~\ref{fig:howtoad}.

Because of the noisy nature of YouTube videos and the abundance of data in the HowTo100M dataset, 
we filter out less preferable videos by the quality of subject detection in HowToCaption, the frequency of names in ASR, and the quality of character portrait faces; details are in the~\app.
As shown in Table~\ref{tab:ad_datasetsl}, the HowTo-AD dataset ends up with a subset of 180k YouTube videos from the original HowTo100M dataset -- which is about 20\% of the full HowTo100M -- and 3.4M transformed AD segments with timestamps from HowToCaption dataset.

\vspace{-1mm}
\section{Model Architecture}
\label{sec:arch}

\begin{figure*}[t]
    \centering
    \includegraphics[width=0.85\textwidth]{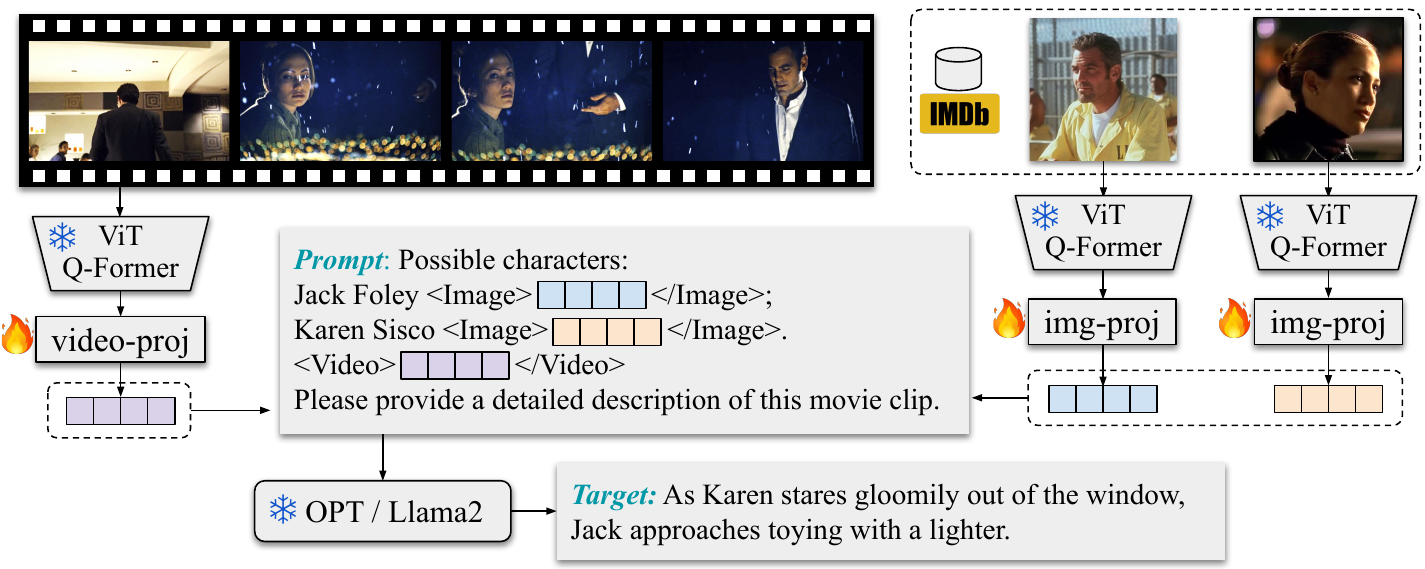}
    \vspace{-0.3cm}
    \caption{\small\textbf{Architecture overview.}
    Our model takes as input movie frames and movie character bank from IMDb including face exemplars and character names, and produces character-aware audio descriptions.
    The input images/videos are first fed to a frozen visual feature extractor to obtain spatial or spatial-temporal visual features.
    Then it uses a shared Q-former to process the visual information and project them to the language embedding space, to leverage frozen large language models(LLM) like OPT and Llama2 for text generation.
    }
    \vspace{-0.5cm}
    \label{fig:arch}
\end{figure*}

With pixel data available,
we propose two visual captioning models based on BLIP2~\cite{li2023blip2} and Llama2~\cite{touvron2023llama2} for movie AD generation.
Specifically, we propose two new architectures called Movie-BLIP2 and Movie-Llama2. Both of them take 8 video frames, resized at $224\times224$ pixels as inputs, 
then use EVA-CLIP~\cite{sun2023evaclip} to extract dense visual features. 
Next, we use a Q-former to attend to spatial-temporal feature grids to extract visual descriptors represented by 32 vectors.
Both models also processes image inputs from character face exemplars. In this case, they take a single image resized at $224\times224$ pixels, and then use the same EVA-CLIP to extract visual features in spatial grid,
and the same Q-former to attend to this spatial feature grid and extract 32 vectors as image descriptors.
The video and image descriptors are passed to two shallow projection heads respectively, to project them on the language embedding space.
Finally, the projected visual outputs together with language prompts are passed to a large language model (OPT for Movie-BLIP2 and Llama2 for Movie-Llama2) to generate movie AD in text form.
An overview of architecture is shown in Figure~\ref{fig:arch}.

The Movie-BLIP2 architecture inherits from the original Image-based BLIP2 architecture, and it uses OPT~\cite{zhang2022opt} as the language model.
The Movie-Llama2 architecture inherits from the image-based MiniGPT-4~\cite{zhu2023minigpt} which connects BLIP2's visual embedding with Llama2 language embedding~\cite{touvron2023llama2}. Our Movie-Llama2 follows the same setup and uses Llama2 as the language model.
We take pre-trained checkpoints from open-sourced projects~\cite{yu2023videoblip} and~\cite{damonlpsg2023videollama}.
Details of these architectures are in the~\app.
Following previous works~\cite{zhu2023minigpt},
by default, all the visual backbone, language model, and the Q-former are frozen, and we only train the projection heads.

\vspace{1mm}
\noindent\textbf{Training details.}
By default, we adopt a two-stage training strategy.
The model is firstly pretrained on HowTo-AD and then finetuned on CMD-AD-Train.
We pretrain on HowTo-AD for 1 epoch and finetune on CMD-AD-Train for 2 epochs. We find finetuning beyond 2 epochs leads to overfitting.
We use a batch size of 8 AD samples, an AdamW optimizer~\cite{loshchilov2017adamw} with $3\times 10^{-5}$ learning rate and a cosine decay schedule.
For both the pretraining and finetuning stages, the training pipeline fits in a single A40 GPU with 48GB GPU memory.
More training details are provided in the~\app.

\vspace{-1mm}
\section{Evaluation Methods}
\label{sec:metrics}
We propose two new methods for evaluating movie AD generation: CRITIC for identifying correct characters, and an LLM-based AD evaluation for assessing holistic semantics of AD.

\vspace{-2mm}
\paragraph{CRITIC (\textbf{C}o-\textbf{R}eferencing \textbf{I}n \textbf{T}ext for \textbf{I}dentifying \textbf{C}haracters).}
The CRITIC metric assesses the accuracy of character naming in predicted AD against human-generated reference AD. The metric is designed to be robust to (i) co-referencing complexities (ii) pronoun usage, and (iii) orthographic variation in character names. 
The objective is to measure the quality of character reference in the generated AD compared to the ground-truth AD. For example, the model might generate AD with the text `Jack' or pronouns like `he', the CRITIC metric aims to evaluate the accuracy of these references.

\begin{figure}[t]
    \centering
    \includegraphics[width=0.49\textwidth]{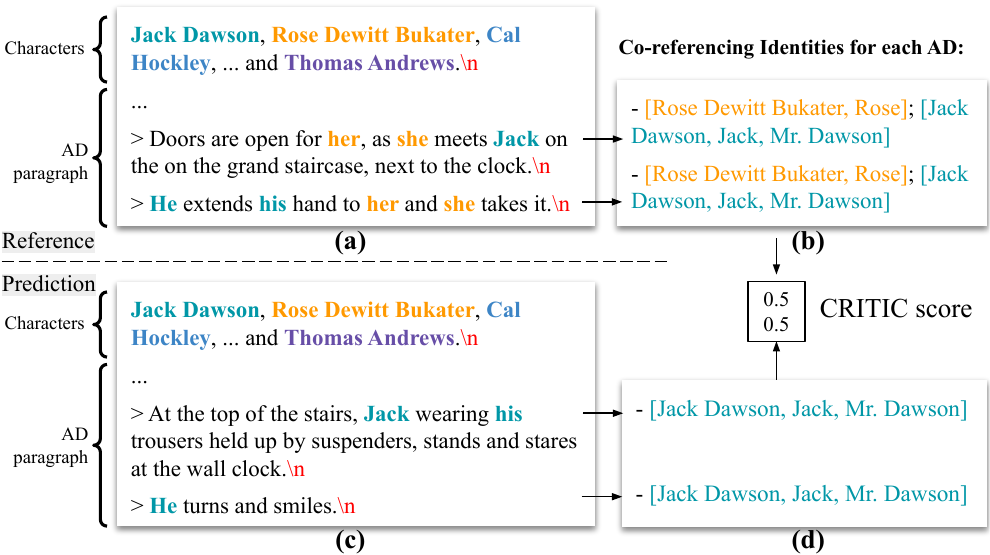}
    \vspace{-0.7cm}
    \caption{\textbf{Illustration of the CRITIC metric.}
    The paragraphs consisting character list and AD (a,c) are fed into a co-referencing model to get co-referencing identities (b,d).
    The CRITIC metric computes an IoU between the identities in the prediction vs.\ the identities from the reference.
    }
    \vspace{-0.5cm}
    \label{fig:critic}
\end{figure}

To achieve this, a \blue{co-referencing} model\footnote{{\scriptsize\url{https://github.com/shon-otmazgin/fastcoref}}} is applied to both predicted and reference AD passages.
Specifically, let $\mathcal{C} = ``c_1, c_2, ..., c_n."$ denote the set of official character names from the cast list of a movie, combined into a single sentence. For a specific movie, we group the predicted and reference audio descriptions into long paragraphs, denoted as $AD_{\text{pred}}$ and $AD_{\text{ref}}$ respectively. 
In order to guide the \blue{co-referencing} model to detect character names,
both $AD_{\text{pred}}$ and $AD_{\text{ref}}$ are prefixed with the character list sentence $\mathcal{C}$, as shown in Figure~\ref{fig:critic} (\textbf{a} and \textbf{c}).

Next, the co-referencing model is applied to both paragraphs from prediction and reference, yielding sets of identities \( E_{\text{pred}} \) and \( E_{\text{ref}} \). Each identity \( E \) includes references and pronouns linked to a single entity.
We only keep identities containing exactly one character name from \( C \), ensuring distinct association with individual characters.
Importantly, we remove pronouns like `he', `she', and `they' in each co-referencing identity to exclude ambiguous pronoun matching.
Next, we map each sentence to its corresponding set of co-referencing identities, which may be empty (when no names are recognized), singular, or multiple, as depicted in Figure~\ref{fig:critic} (\textbf{b} and \textbf{d}).

The CRITIC metric $M_{\text{CRITIC}}$ is then calculated as \blue{an IoU},
for $i$-th AD reference (with valid character identities):
\vspace{-1mm}
\begin{equation}
    M_{\text{CRITIC}} = \frac{|E_{\text{pred}} \cap E_{\text{ref}}|}
    {|E_{\text{pred}} \cup E_{\text{ref}}|}
\end{equation}
\vspace{-3mm}

\noindent where $|E_{\text{pred}} \cap E_{\text{ref}}|$ is the count of matching identities between predicted and reference ADs, and $|E_{\text{pred}} \cup E_{\text{ref}}|$ is the total count of unique identities in \blue{both the prediction and the reference}. The CRITIC score is averaged across all the AD references.
Intuitively, if the predicted AD includes a name, the CRITIC score verifies whether the name refers to the correct identity; if the prediction includes a pronoun like `he', the CRITIC score first resolves the identity by co-referencing, then verifies whether the name is correct.
The CRITIC score has a range between 0 and 1, with 1 being perfect.

\vspace{-2mm}
\paragraph{LLM-AD-eval.}
We also adopt the LLM as a judge~\cite{zheng2023judging, chiang2023can} procedure for AD quality assessment.
Following previous works~\cite{maaz2023video}, 
we use a `gpt-3.5-turbo' API from OpenAI
\blue{and an open-sourced `llama-2-7b-chat' model~\cite{touvron2023llama2}.}
prompting the model to assess the matching quality between a pair of predicted AD and ground-truth AD by a score from 1 to 5, where 5 indicates the best matching and 1 indicates the worst matching.
To be complementary to the CRITIC metric, 
for LLM-AD-eval we instruct the LLM to (1) consider pronouns as valid matches and ignore character names, and (2) focus on human actions, objects and interactions.
The customized prompts are provided in the~\app.

%% file: sec/4_exp.tex
\section{Experiments}
\label{sec:exp}
We outline the datasets (Sec.~\ref{subsec:exp:data}) and evaluation measures (Sec.~\ref{subsec:exp:eval})
employed in our experiments, and
provide an analysis on inter-rater agreement between AD annotation versions
(Sec.~\ref{subsec:exp:interrater}).
We report quantitative results, ablating the architectural design and the effect of HowToAD pretraining (Sec.~\ref{subsec:exp:quantitative}),
followed by
qualitative results thanks to our pixel movie data (Sec.~\ref{subsec:exp:qual}).

\subsection{Datasets}
\label{subsec:exp:data}
\textbf{AudioVault-8k} is the dataset collected from~\url{https://audiovault.net/} by~\cite{Han23a} that contains full-movie AD and subtitles transcribed from user-uploaded audio description files covering 7800 movies. 
\textbf{CMD (Condensed Movie Dataset)}~\cite{Bain20} contains movie clips collected from YouTube for more than 3k movies. On average each movie has 10 non-contiguous clips and each clip spans for a few minutes. 
\textbf{HowTo100M}~\cite{miech2019howto100m} contains 1M YouTube long videos with more than 100M ASR segments. It is typically used for video pretraining.
\textbf{CMD-AD} is the new movie AD dataset introduced in Section~\ref{subsec:data:cmd}, by aligning AD data with CMD clips~\cite{Bain20}. It contains 101k ADs for more than 1432 movies. We split a 100-movie evaluation set named CMD-AD-Eval and use the rest for training.
\textbf{HowTo-AD} is the new AD dataset introduced in Section~\ref{subsec:data:howtoad}, transformed from HowTo100M~\cite{miech2019howto100m}. It contains 180k YouTube videos with augmented descriptions and character exemplars. We mainly use it for AD generation pertaining.

\vspace{-1mm}
\subsection{Evaluation Measures}
\label{subsec:exp:eval}

In addition to the two new evaluation measures introduced in Section~\ref{sec:metrics}, \textbf{CRITIC} and \textbf{LLM-AD-Eval}, we also monitor Recall@k/N, CIDEr, and perplexity.
\textbf{Recall@k/N}~\cite{Han23a} is a retrieval metric that distinguishes the predicted text among a set of temporal neighbours.
The parameters $k$ and $N$ mean within a temporal window of $N$ neighbouring reference ADs, 
whether the predicted AD can retrieve the corresponding reference AD at top-$k$ position. We use Recall@1/5 on CMD-AD-Eval and Recall@5/16 on MAD-Eval to compare with previous works. We use the official implementation provided by~\cite{Han23a}.
\textbf{CIDEr}~\cite{vedantam2015cider} is a popular text similarity metric that is based on word matching rate.
We include Recall@k/N and CIDEr here as they have been used in recent work on AD~\cite{Han23,Han23a} and we also compare on the test datasets of those works.

\begin{figure}
    \centering
    \includegraphics[width=0.48\textwidth]{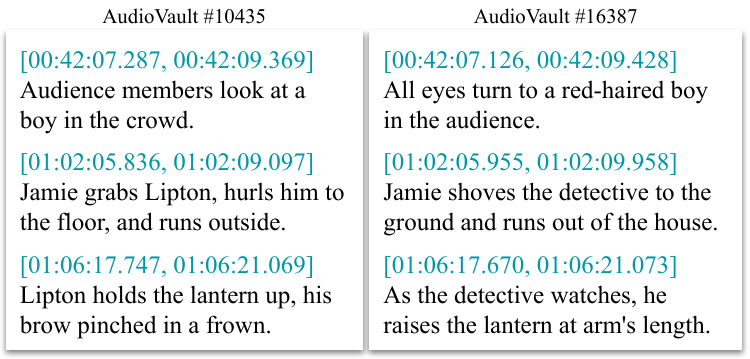}
    \vspace{-0.7cm}
    \caption{\textbf{An example of inter-rater evaluation.}
    Some movies on AudioVault have multiple available ADs. Two versions of human-annotated ADs for the same visual scene are shown here. These ADs are filtered with a tIoU threshold of 0.9, and from the movie `Dead Silence'(\texttt{tt0455760}).
    }
    \vspace{-0.1cm}
    \label{fig:inter_rater}
\end{figure}

\begin{table} 
    \centering
    \resizebox{0.99\linewidth}{!}{
    \setlength{\tabcolsep}{3pt}
    \begin{tabular}{ccc|cccc}
    \toprule
        tIoU & \#movies & \#AD pairs & CIDEr & R@1/5 & CRITIC & LLM-AD-eval$\dagger$ \\ \midrule
        0.8 & \blue{315}        & \blue{4447}  & 61.5  & 71.2   & \blue{42.0}  & 2.56 / \blue{3.04}  \\
        0.9 & \blue{267}        & \blue{999}   & 69.8  & 80.4   & \blue{47.6}  & 3.06 / \blue{3.53}  \\
    \bottomrule
    \end{tabular}
    }
    \vspace{-0.3cm}
    \caption{\textbf{Inter-rater agreement on AudioVault AD annotations.}
    AD from different annotators do not usually synchronize. 
    A higher tIoU threshold filters out fewer AD pairs but they are more likely to describe the exact same visual event.
    \blue{$\dagger$:~LLM-AD-Eval scores are computed from `gpt-3.5-turbo' / `llama-2-7b-chat' respectively.}
    For all the metrics, a higher number indicates better quality. R@1/5 and CRITIC are upperbounded at 100 and LLM-AD-eval is between 1 to 5.
    }
    \vspace{-0.5cm}
    \label{tab:interrater}
\end{table}

\subsection{Inter-rater Evaluations}
\label{subsec:exp:interrater}

Many of the films in Audiovault have multiple ADs available. Typically these are UK and US versions. In this section, we use the agreement between the human-provided AD versions 
to assess the usefulness of the four evaluation metrics -- CIDEr, Recall@1/5, CRITIC, and LLM-AD-eval. Note, these evaluations are carried out directly on the text version of the AD, so no pixels are involved. 

In the AudioVault-8K dataset, 
there are 402 movies with more than one version of AD annotations.
We conduct inter-rater experiments on this subset of AD annotations.
Three challenges emerge when conducting inter-rater comparisons for the same visual scene:
(i) Different versions of AD annotations might correspond to different versions of the same movie which do not naturally synchronize as introduced in Section~\ref{subsec:data:cmd}. We apply the audio-audio alignment pipeline in~\ref{subsec:data:cmd} to synchronize both AD annotations.
(ii) The timing of providing AD is subjective and arbitrary within a short time interval~\cite{Han23a}, to obtain different ADs for the exact same visual moment, we have to filter the time segments of two AD versions with a temporal Intersection-over-Union (tIoU).
(iii) For about 20\% of movies, the multiple AD versions from AudioVault are simply narrating the same scripts again with minor modifications, 
which does not reflect independent inter-rater comparisons. We filter out those movies by checking the exact sentence-matching rate.

\begin{table} 
    \centering
    \resizebox{1\linewidth}{!}{
    \setlength{\tabcolsep}{3pt}
    \begin{tabular}{lll|cccc}
    \toprule
        Method     & V-model       & L-model          & CIDEr & R@1/5 & CRITIC & LLM-AD-eval$\dagger$  \\ \midrule
        AutoAD-II   & CLIP-B-32     & GPT2                  & 13.5   & 26.1  & \blue{8.2}  & 1.53 / \blue{2.08}      \\ 
        Movie-BLIP2 & Eva-CLIP      & OPT-2.7B              & 21.2  & 29.3  & \blue{{24.5}}  &  \textbf{2.13} / \blue{2.66}       \\
        Movie-Llama2& Eva-CLIP      & LLama2-7B             & \textbf{21.7}  & \textbf{30.0 }& \blue{\textbf{25.2}}  &  2.05 / \blue{\textbf{2.85}}       \\   
    \bottomrule
    \end{tabular}
    }
    \vspace{-3mm}
    \caption{\textbf{Architecture experiments} on CMD-AD-Eval.
    We compare the proposed two architectures with AutoAD-II. All of them take character bank inputs. 
    $\dagger$:~from `gpt-3.5-turbo' / `llama-2-7b-chat' respectively.
    Note, AutoAD-II is trained with averaged frame features to mimic its original setting on the feature-only MAD dataset.
    }
    \vspace{-1mm}
    \label{tab:arch}
\end{table}

The inter-rater evaluation is shown in Table~\ref{tab:interrater}.
With a higher tIoU threshold, we get fewer AD annotation pairs covering fewer movies, but the AD annotation pairs are more likely to describe the same movie scene.
The numbers can be regarded as human-level upperbound.
Note that with a lower tIoU (from 0.9 to 0.8), all the metrics drop significantly, highlighting the temporal sensitivity of AD tasks and the importance of precise data alignment.
A few pairs of human-annotated AD from AudioVault are shown in Figure~\ref{fig:inter_rater}, where the left and right panels are from two annotators for the same visual scene.

\begin{figure*} 
    \centering
    \includegraphics[width=0.98\textwidth]{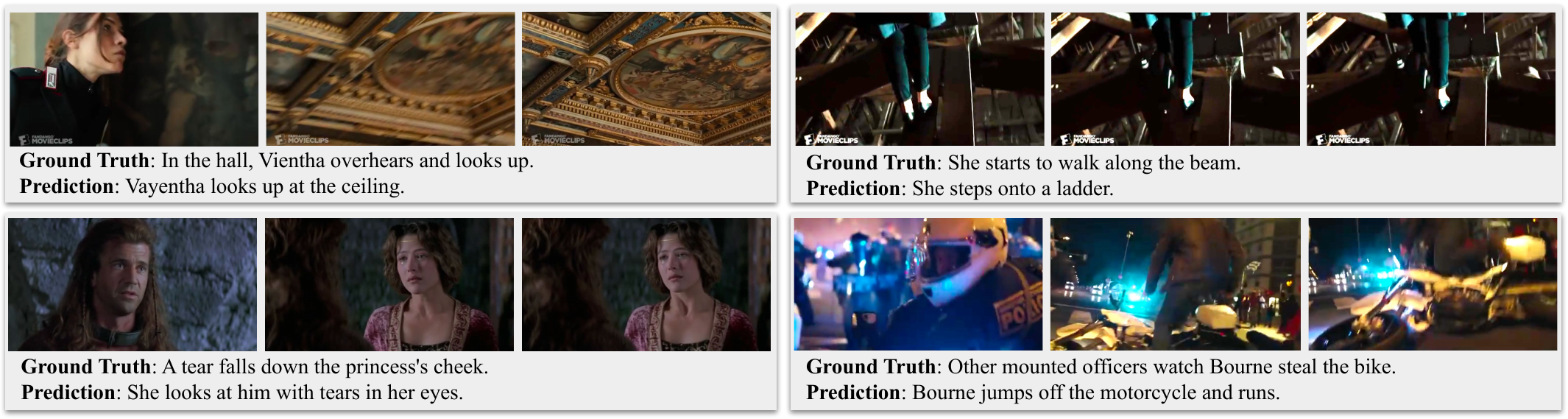}
    \vspace{-0.3cm}
    \caption{\textbf{Qualitative results.} AutoAD-III predictions
    correctly identify the semantics of the scene, by referring to the characters (`Vayentha', `Bourne'),
    their relations (`looks at him'), actions (`steps onto', `jumps off'), emotions (`tears'), objects (`ladder', `motorcycle'). The comparison with the ground truth further highlights the limitations of the n-gram based metrics since the same meaning can be conveyed with different wordings.
    }
    \vspace{-0.4cm}
    \label{fig:qualitative}
\end{figure*}

\begin{table} 
    \centering
    \resizebox{1\linewidth}{!}{
    \setlength{\tabcolsep}{4pt}
    \begin{tabular}{lc|cccc}
    \toprule
        Method       & pretrain & CIDEr  & R@1/5 & CRITIC & LLM-AD-eval$\dagger$ \\
        \midrule
        Movie-BLIP2  & \xmark       & 21.2   & 29.3  & \blue{24.5}   & 2.13 / \blue{2.66}     \\
        Movie-BLIP2  & HowTo-AD      & \textbf{22.3} & \textbf{29.8}   & \blue{\textbf{30.2}}   &     \textbf{2.25} / \blue{\textbf{2.78}}  \\
        \midrule
        Movie-Llama2  & \xmark       & 21.7   & 30.0  & \blue{25.2}   &  2.05 / \blue{2.85}        \\
        Movie-Llama2  & HowToCaption$\ddagger$ & 20.8  & 29.4   & \blue{25.6}      &  2.07 / \blue{2.85}  \\
        Movie-Llama2  & HowTo-AD      & \textbf{25.0}   & \textbf{31.2}  & \blue{\textbf{32.7}}   &   \textbf{2.29} / \blue{\textbf{2.92}}     \\
    \bottomrule
    \end{tabular}
    }
    \vspace{-0.2cm}
    \caption{\textbf{Effect of HowTo-AD pretraining}
    on CMD-AD-Eval. 
    $\dagger$:~from `gpt-3.5-turbo' / `llama-2-7b-chat', respectively.
    $\ddagger$~uses the same 180k-video subset as HowTo-AD, but without constructing character banks or rewriting captions.
    }
    \vspace{-0.4cm}
    \label{tab:result_howtoad}
\end{table}

\begin{table}
    \centering
    \resizebox{1\linewidth}{!}{
    \setlength{\tabcolsep}{3pt}
    \begin{tabular}{l|cccc|cc}
    \toprule
        \multirow{2}{*}{Method} & \multicolumn{4}{c|}{CMD-AD-Eval}                                                                                      & \multicolumn{2}{c}{MAD-Eval} \\
                                & \multicolumn{1}{l}{CIDEr} & \multicolumn{1}{l}{R@1/5} & \multicolumn{1}{l}{CRITIC} & \multicolumn{1}{l|}{LLM-AD-Eval$\dagger$} & CIDEr         & R@5/16        \\ \midrule
        Video-BLIP2~\cite{yu2023videoblip} (no ft)  & 4.8                       &  22.0                         & \blue{0.0}                           &  1.40 / \blue{1.89}                                &  \textcolor{white}{0}5.0\textcolor{white}{*}             & 35.2\textcolor{white}{*}             \\
        Video-Llama2~\cite{damonlpsg2023videollama} (no ft) & 5.2                       &  23.6                         &  \blue{0.0}                          & 1.43 / \blue{1.91}                             &  \textcolor{white}{0}4.8\textcolor{white}{0}             &  33.8\textcolor{white}{*}            \\ \midrule
        
        \blue{MM-Narrator-GPT4~\cite{zhang2023mm}} & - & - & - & - & \blue{13.9\textcolor{white}{*}} & - \\
        
        AutoAD-I~\cite{Han23}               & -                       & -                      & -                       & -                             & 14.3\textcolor{white}{*}           & 42.1\textcolor{white}{*}         \\
        AutoAD-II~\cite{Han23a}               & 13.5                       & 26.1                      & \blue{8.2}                       & 1.53 / 2.08                             & 19.2\textcolor{white}{*}          & 51.3\textcolor{white}{*}         \\
        Movie-BLIP2 \textbf{(ours)}     & 22.3                      & 29.8                     &  \blue{30.2}                       & 2.25 / 2.78                    & 22.8*          & 52.0*        \\
        Movie-Llama2 \textbf{(ours)}     & 25.0                      & 31.2                     &  \blue{32.7}                       & 2.29 / 2.92                    & 24.0*          & 52.8*        \\
    \bottomrule
    \end{tabular}
    }
    \vspace{-0.2cm}
    \caption{
    \textbf{Comparison with other methods on CMD-AD-Eval and MAD-Eval.}
    Additionally, we evaluate the out-of-the-box Video-BLIP2 and Video-Llama2 video captioning models (without any AD finetuning) directly on both datasets. 
    $\dagger$: from `gpt-3.5-turbo' / `llama-2-7b-chat' respectively.
    *: these results are obtained on MAD-Eval \emph{without} any training on MAD-Train. 
    }
    \vspace{-0.4cm}
    \label{tab:sota}
\end{table}

\vspace{-1mm}
\subsection{Quantitative Results}
\label{subsec:exp:quantitative}

\paragraph{Architecture Comparisons on Aligned-CMD.}
In Table~\ref{tab:arch}, 
we compare the proposed Movie-BLIP2 and Movie-Llama2 architectures
with previous methods on the CMD-AD dataset.
All of these models are trained on the CMD-AD-Train set and evaluated on the CMD-AD-Eval set. 
We implement and train AutoAD-II on CMD-AD-Train based on the public codebase.
The results show that both Movie-BLIP2 and Movie-Llama2 perform much better than AutoAD-II architecture. The stronger performance is attributed to multiple factors -- stronger visual backbone, stronger language model, and taking visual \emph{grid} feature as input instead of a single vector as in AutoAD-II.
Movie-Llama2 has a much stronger language model than Movie-BLIP2 (Llama2-7B vs OPT-2.7B), but it achieves a similar performance wrt Movie-BLIP2.
Note that all these models are not pretrained on HowTo-AD yet.

\vspace{-3mm}
\paragraph{Effect of HowTo-AD Pretraining.}
Taking the Movie-BLIP2 and Movie-Llama2 settings from Table~\ref{tab:arch},
we compare the effect of HowTo-AD by pretraining the same architecture on the HowTo-AD dataset then finetuning on CMD-AD-Train set.
We also pretrain with the same subset from HowToCaption without using the character bank or rewriting captions.
The results in Table~\ref{tab:result_howtoad} show that large-scale pretraining on our HowTo-AD dataset substantially boosts the performance on all four metrics for both models.~\eg~improving CRITIC from \blue{25.2 to 32.7} and CIDEr from 21.7 to 25.0 for Movie-Llama2.
But pretraining on HowToCaption does not help much on the finetuned movie AD task, possibly because of the domain gap from the data and task.

\vspace{-3mm}
\paragraph{Comparison with Other Methods.}
We compare with other methods on two datasets:
CMD-AD-Eval introduced in this work,
and MAD-Eval proposed in~\cite{Han23} (Table~\ref{tab:sota}).
Note that MAD-Eval is a 10-movie subset from LSMDC, where we can get short movie clips for evaluation. However, we can not perform any training on MAD-Train since no pixels are available.
The proposed method Movie-BLIP2 and Movie-Llama2 perform much better than the previous methods including MM-Narrator with GPT4 as the language model.
We also evaluate video captioning models like Video-BLIP2 and Video-Llama2 but neither of them performs well on AD, highlighting the challenges of Movie AD task.

\subsection{Qualitative Analysis}
\label{subsec:exp:qual}
Figure~\ref{fig:qualitative} illustrates several random examples from the CMD-AD-Eval set.
For each sample, we display the predictions of our Movie-Llama2
model, as well as the ground truth AD. 
We observe that, while different wording than the ground truth, the semantics of the AD content remain largely similar for Our method.
Interestingly, in the first example, the ASR pipeline~\cite{Bain23} transcribed the name incorrectly as `Vientha' but our model fixed the name through the character bank.
More examples can be found in the~\app.

%% file: sec/9_conclusion.tex
\vspace{-1mm}
\section{Conclusion}
\label{sec:conclusion}

This work advances automatic AD generation for movies by:
(i)~collecting AD for pixel data through audio-audio alignment between full movies (without pixels) and public movie snippets, and pseudo-labelling instruction videos;
(ii)~showing that recent video-language architectures provide a significant performance boost, bringing AD generation systems closer to real-world applications;
and (iii)~proposing new evaluation methods tailored for AD.
One of the limitations that necessitates future work is the coherence
across AD narrations throughout the movie: AD should not repeat the same information,
or provide story-irrelevant details. To this end, external knowledge such as plot summaries
may be utilized to incorporate story-centric elements.
Future directions could also explore the
harmony between the narration tone and the movie content
for an engaging experience.

\vspace{3mm}
\blue{
\noindent\textbf{Acknowledgements.}
This research is funded by EPSRC PG VisualAI EP/T028572/1, 
and ANR-21-CE23-0003-01 CorVis.
}

%% file: sec/X_appendix.tex
\clearpage
\appendix


\renewcommand{\thefigure}{A.\arabic{figure}} 
\setcounter{figure}{0} 
\renewcommand{\thetable}{A.\arabic{table}}
\setcounter{table}{0} 

\twocolumn[{%
\renewcommand\twocolumn[1][]{#1}%
\maketitleappendix
\setlength{\tabcolsep}{2pt}
\begin{center}
    \centering
\vspace{-7mm}
\begin{tabular}{c}
\includegraphics[width=1.0\textwidth]{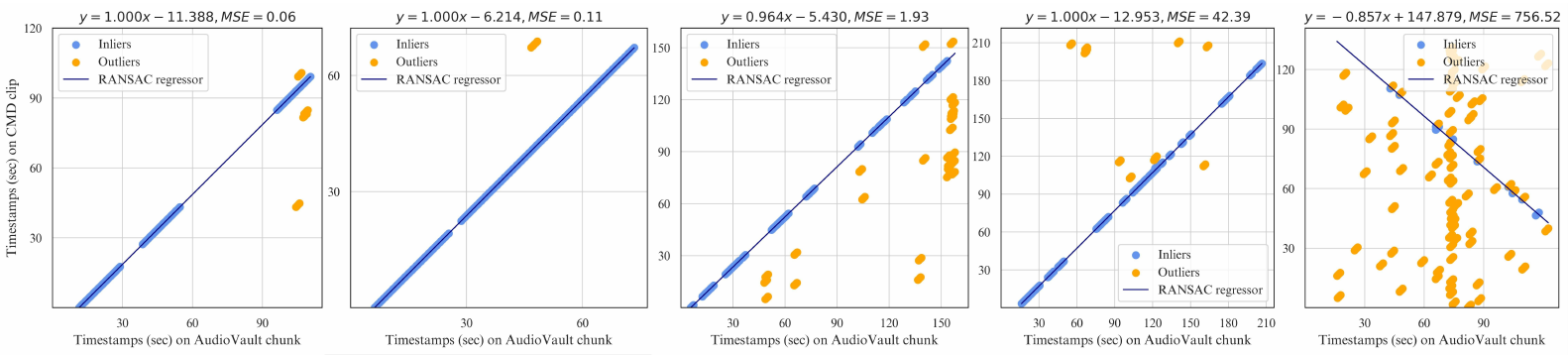}
\label{fig:supp:ransac}\\ \vspace{-0.1cm}(a)\\
\includegraphics[width=1.0\textwidth]{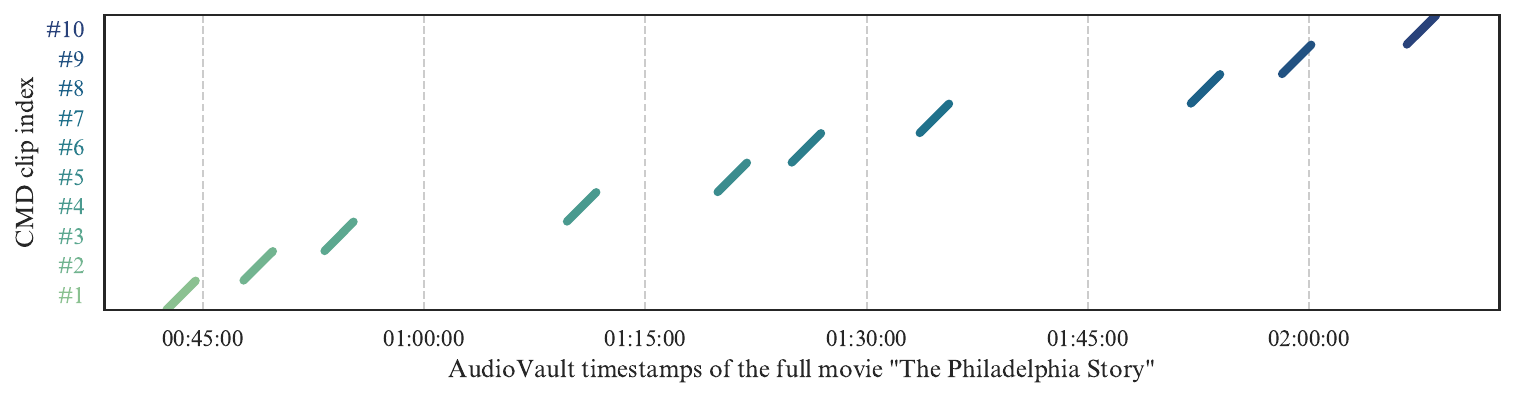}
\label{fig:supp:fullmovie_aa}\\ \vspace{-0.2cm}(b)
\end{tabular}
\captionof{figure}{\textbf{(a)} More examples of RANSAC for audio-audio alignment.
    Five samples are chosen from five different movies.
    The four samples on the left are successful RANSAC results that have low MSE and reasonable slopes. The sample on the right is filtered out since the fitted result from RANSAC has a high MSE and the slope does match our criteria.
    Note that the inliers  (\textcolor[HTML]{6794E7}{blue} dots) are not continuous because those time segments correspond to AD audio and have been masked out before the alignment process.
    \textbf{(b)} Audio-audio alignment results of a full movie.
    We localize ten CMD clips from the corresponding \emph{full-movie} AudioVault audio file by visualizing their aligned timestamps from the RANSAC algorithm. We can clearly see that CMD dataset contains multiple discontinuous movie clips, but the precise timestamps of each clip are obtained from our alignment algorithm, which are not provided by CMD.
    The example movie is `The Philadelphia Story' (1940, \texttt{tt0032904}).
    }
\label{fig:supp:ransac_pair}
\end{center}%
}]

\appendix
\vspace{25mm}
\section{Dataset Details}

\subsection{CMD-AD -- Pixels from Aligned CMD}
Following the main paper Figure~2,
a few more examples of RANSAC output are shown in Figure~\ref{fig:supp:ransac_pair}-(a).
Note that we filter valid RANSAC output by two thresholds: (1) the slope $0.8<W'<1.25$, and (2) MSE$<100$. 
Additionally, we visualize the audio-audio alignment results of a \emph{full movie} example in Figure~\ref{fig:supp:ransac_pair}-(b). 

The script for conducting audio-audio alignment between CMD movie clip and AudioVault audio chunk is shown in Algorithm~\ref{code:aa_match}.

\subsection{HowTo-AD -- Pixels from HowTo100M}

The original HowTo100M and HowToCaption datasets have more than 1.2M videos.
We apply a sequence of filtering processes to get high-quality videos for our purpose.
(i) we use the \texttt{spacy} package in Python to detect unique \emph{names} in the YouTube ASR. We discard videos with more than 5 unique names, because we find those videos typically correspond to News or Sports programmes and are not suitable to be transformed to movie AD style.
(ii) we use the \texttt{spacy} package to detect the \emph{subject} of each sentence in the HowToCaption dataset. We discard those videos which do not contain a subject in any of the captions,~\eg the captions like `cut carrots', `chunk onion',~\etc. Because our caption transforming stage in the main paper Section 3.2 requires the detected subjects.
(iii) we extract a few video frames at the introduction of the video -- which is inferred from the first few ASR timestamps -- then we use a face detector\footnote{\url{https://github.com/ageitgey/face_recognition}} to verify whether a face exists in these frames. We discard videos that do not have frames with faces because our character bank requires face exemplars.

A combination of these filters reduces the number of videos from 1.2M to 180k, from which we build the HowTo-AD dataset as described in the main paper Section 3.2.

\begin{figure}[t]
    \centering
    \includegraphics[width=0.49\textwidth]{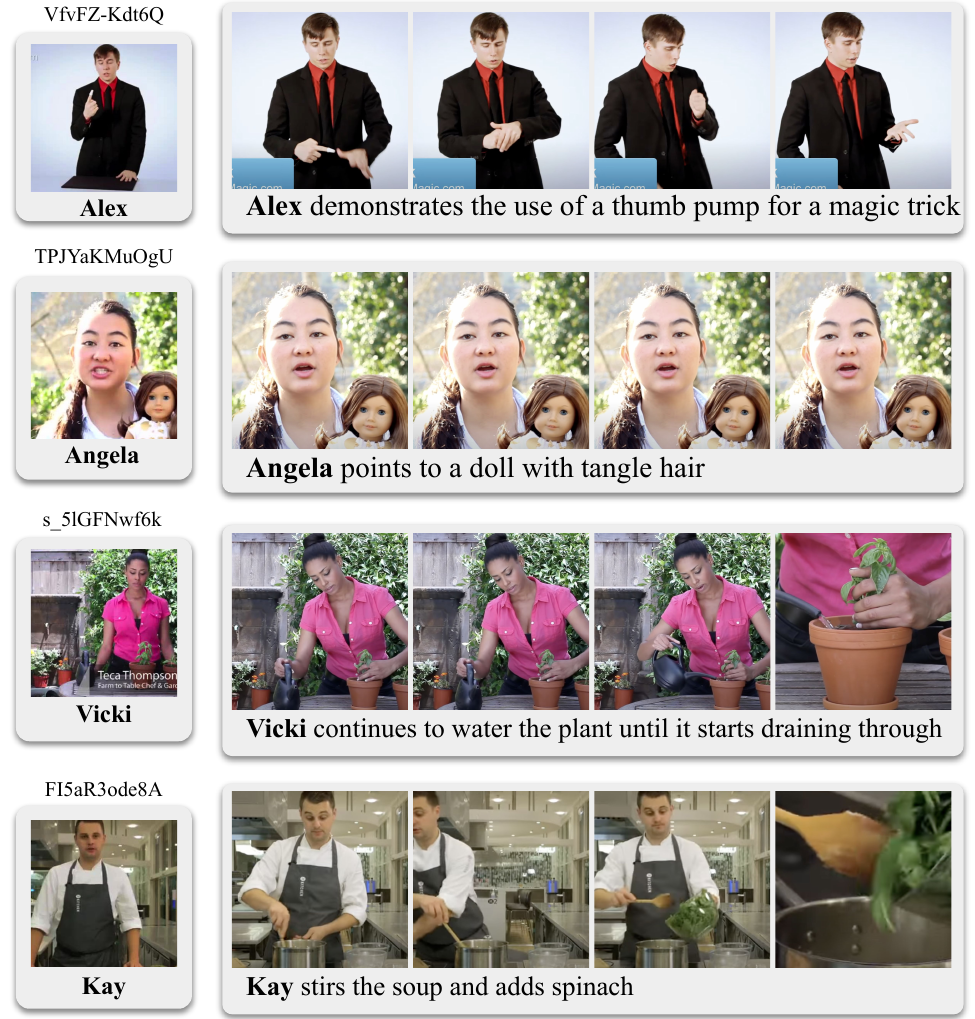}
    \caption{More samples from HowTo-AD dataset.
    Four examples are shown in this figure with the `character bank' shown on the left and the re-written video caption with randomly sampled character names shown on the right.
    The YouTube IDs are displayed on top of the character banks.}
    \label{fig:supp:howtoad}
\end{figure}

\begin{figure}[h!]
    \centering
    \includegraphics[width=0.48\textwidth]{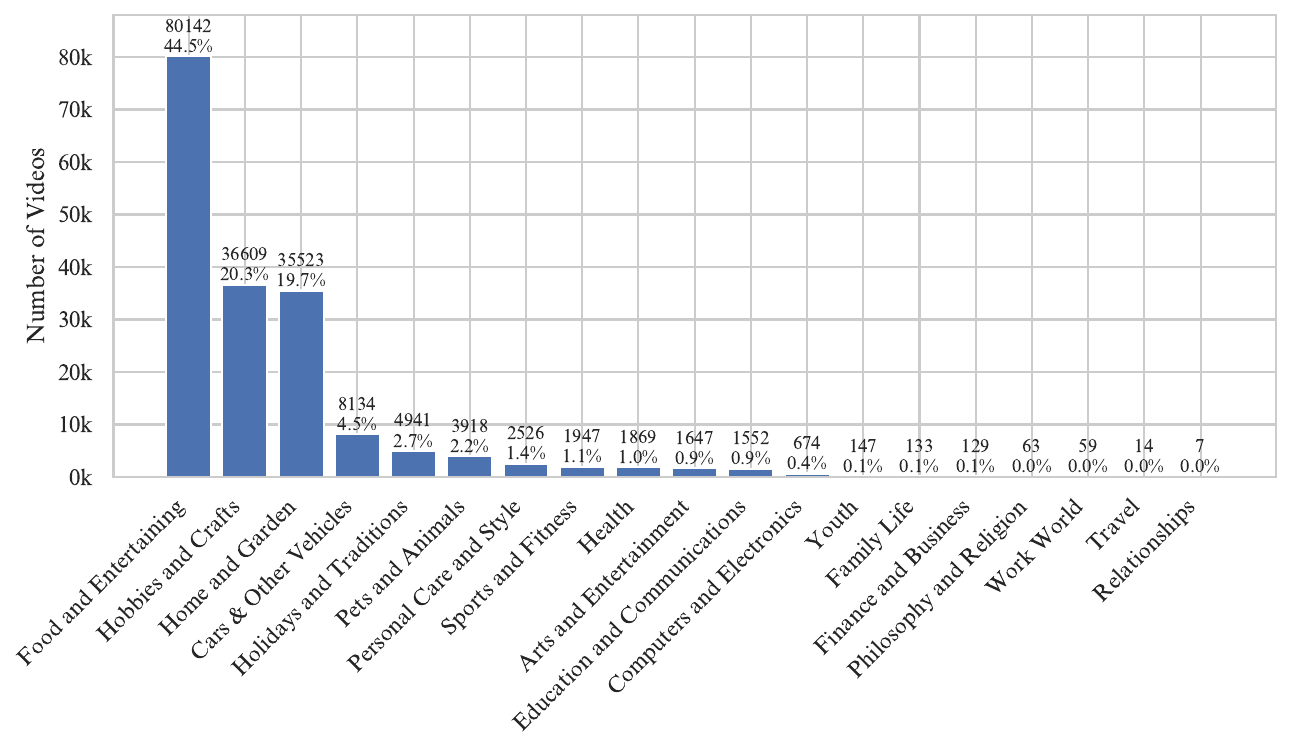}
    \caption{Category distribution of HowTo-AD dataset.
    Similar to the original HowTo100M dataset, the HowTo-AD dataset has a long-tailed distribution over action categories, dominated by `Food and Entertaining', `Hobbies and Crafts' and `Home and Garden'.
    }
    \label{fig:supp:howtoad_category}
\end{figure}

In Figure~\ref{fig:supp:howtoad}, we show a few more examples from HowTo-AD dataset.
We notice a few limitations of the HowTo-AD: 
(1) inevitably, the video categories are dominated by `Food and Entertaining', in other words, cooking activities. Since the dataset originates from HowTo100M where 50\% of the categories belongs to cooking activities. A breakdown of category statistics is shown in Figure~\ref{fig:supp:howtoad_category}.
(2) our current pipeline that transforms HowToCaption into HowTo-AD assumes the video only has a single character. However, it's common that in instructional videos, two or three instructors appear in the scene. The single-character assumption may introduce noisy AD in the HowTo-AD dataset, and we consider fixing this in future work.

\section{Implementation Details}

\subsection{Architectures}
\paragraph{Movie-Llama2} is based on the visual branch of the Video-Llama~\cite{damonlpsg2023videollama} repository\footnote{\url{https://github.com/DAMO-NLP-SG/Video-LLaMA}}.
The input to the model is 8 movie frames uniformly sampled from a given AD time segment. 
The model contains five components, as detailed below.
(i) The visual encoder is from EVA-CLIP which takes input from $224\times224\times3$ pixels and uses $14\times14$ patch size.
The architecture contains 39 transformer blocks with a consistent hidden dimension of 1408 channels and 16 heads.
(ii) The Q-former is from BLIP-2, which contains 12 transformer blocks with 32 learnable queries and cross-attends visual features with a dimension of $(1+16^2)\times768$. For each image input, it produces features with a dimension of $32\times768$.
(iii) The video Q-Former is from Video-Llama, which also contains 12 transformer blocks with 32 learnable queries. It cross-attends video features with a dimension of $8\times 32 \times 768$ and produces $32 \times 768$ video features.
(iv) The projector is a linear layer that projects  $32 \times 768$ visual features into $32 \times 4096$ dimension to fit the language model.
We have two projectors for video and image inputs respectively.
(v) The language model is Llama2-7B, which contains 32 transformer blocks with $4096$ hidden dimensions.

By default, we only train the parameters in the video Q-Former described in (iii) and the projector described in (iv).
The rest of the components are initialized from Video-Llama pretrained weights.
We also experimented with a setting that only trains the linear projector (iv), it produces a lower performance (CIDEr drops by 1.5).

\paragraph{Movie-BLIP2} is based on the Video-BLIP~\cite{yu2023videoblip} repository\footnote{\url{https://github.com/yukw777/VideoBLIP}}.
With an input of 8 movie frames, 
the architecture is mostly the same as Video-Llama, except for two differences. 
First, the functions of the Q-former and video Q-Former are merged. In Movie-BLIP2, the Q-former cross-attends to \emph{all} the visual features at once with a dimension of $8\times(1+16^2)\times1408$, then produces video features $32\times768$. There is no \emph{video} Q-Former in Movie-BLIP2.
Second, the language model is OPT-2.7B, which contains 32 transformer blocks with $2560$ hidden dimensions.

Similarly, we train the parameters in the Q-former and the projector. 
The rest of the components are initialized from Movie-BLIP pretrained weights. But only training the projector produces a similar performance.


\subsection{LLM-AD-Eval}

We use the \texttt{gpt-3.5-turbo} model from OpenAI to construct the LLM-AD-Eval metric. Our customized prompt is shown in Algorithm~\ref{code:llm_ad_eval}.
The LLM is instructed to evaluate the level of match between two sentences and return a score between 0 to 5.

\section{Additional Results}

\subsection{Details of Inter-rater Analysis on AudioVault}

\begin{figure}[t]
    \centering
    \includegraphics[width=0.48\textwidth]{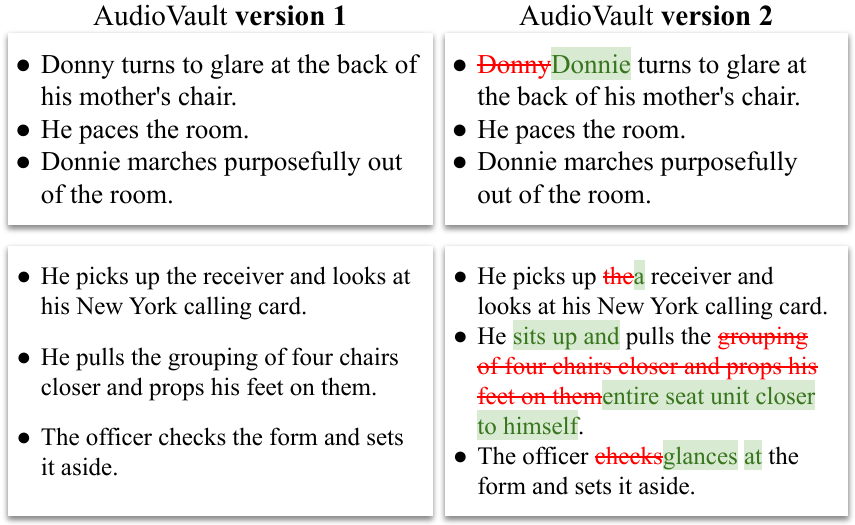}
    \caption{Special cases of inter-rater analysis:
    Two versions of AD from AudioVault might not be produced independently.
    In these two examples, annotators edit a few words from the earlier version, to produce a new version of AD.
    We \emph{discard} these samples when conducting inter-rater analysis since these examples do not reflect real human-human agreement on AD.
    }
    \label{fig:supp:inter_rater}
\end{figure}

\begin{figure}[t]
    \centering
    \includegraphics[width=0.5\textwidth]{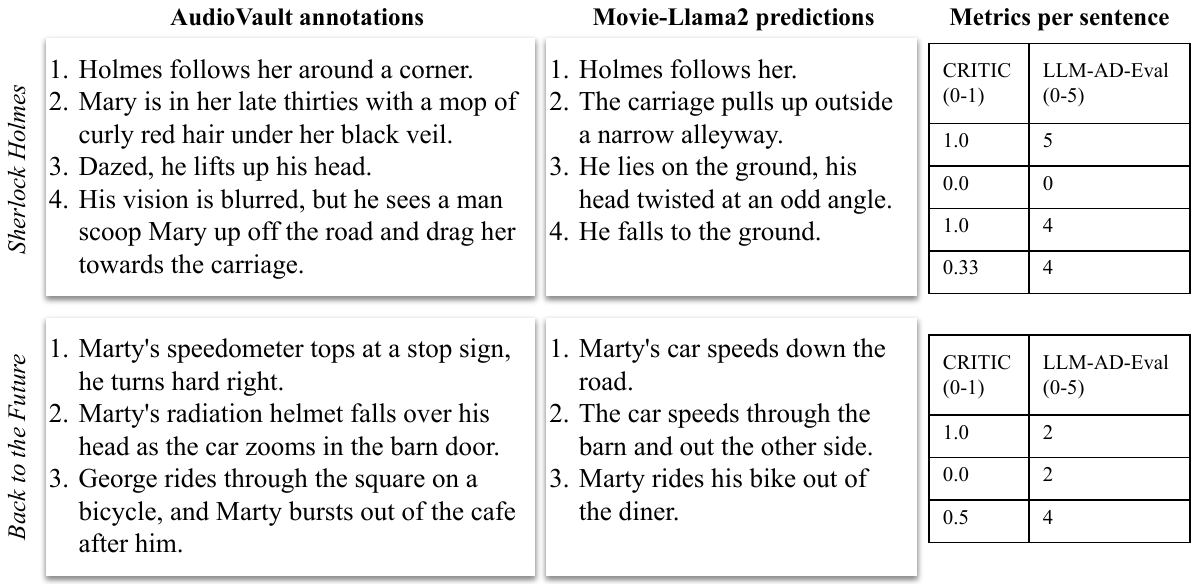}
    \caption{Examples of CRITIC and LLM-AD-Eval metrics.
    We show seven AD annotation-prediction pairs from two movies.
    For both CRITIC and LLM-AD-Eval metrics, higher values indicate higher quality.
    For details of metrics please refer to the main paper Section 5.
    }
    \label{fig:supp:metrics}
\end{figure}

As described in the main paper Section 6.3,
there are 402 movies with multiple versions of AD annotations.
However, we found in some cases two versions of AD annotations are not produced \emph{independently}, as shown in Figure~\ref{fig:supp:inter_rater} -- annotators slightly modify an older version of AD to produce a new version, therefore these samples do not reflect the real human-human agreement on AD annotations.
To discard these samples in our inter-rater analysis,
we apply a filter to detect the identical AD annotations for each movie and discard those movies with more than one identical annotations.
It ends up filtering out $18\%$ from 402 movies, and the remaining trustworthy independent AD annotations are used for inter-rater analysis.

\begin{figure}[t]
    \centering
    \includegraphics[width=0.48\textwidth]{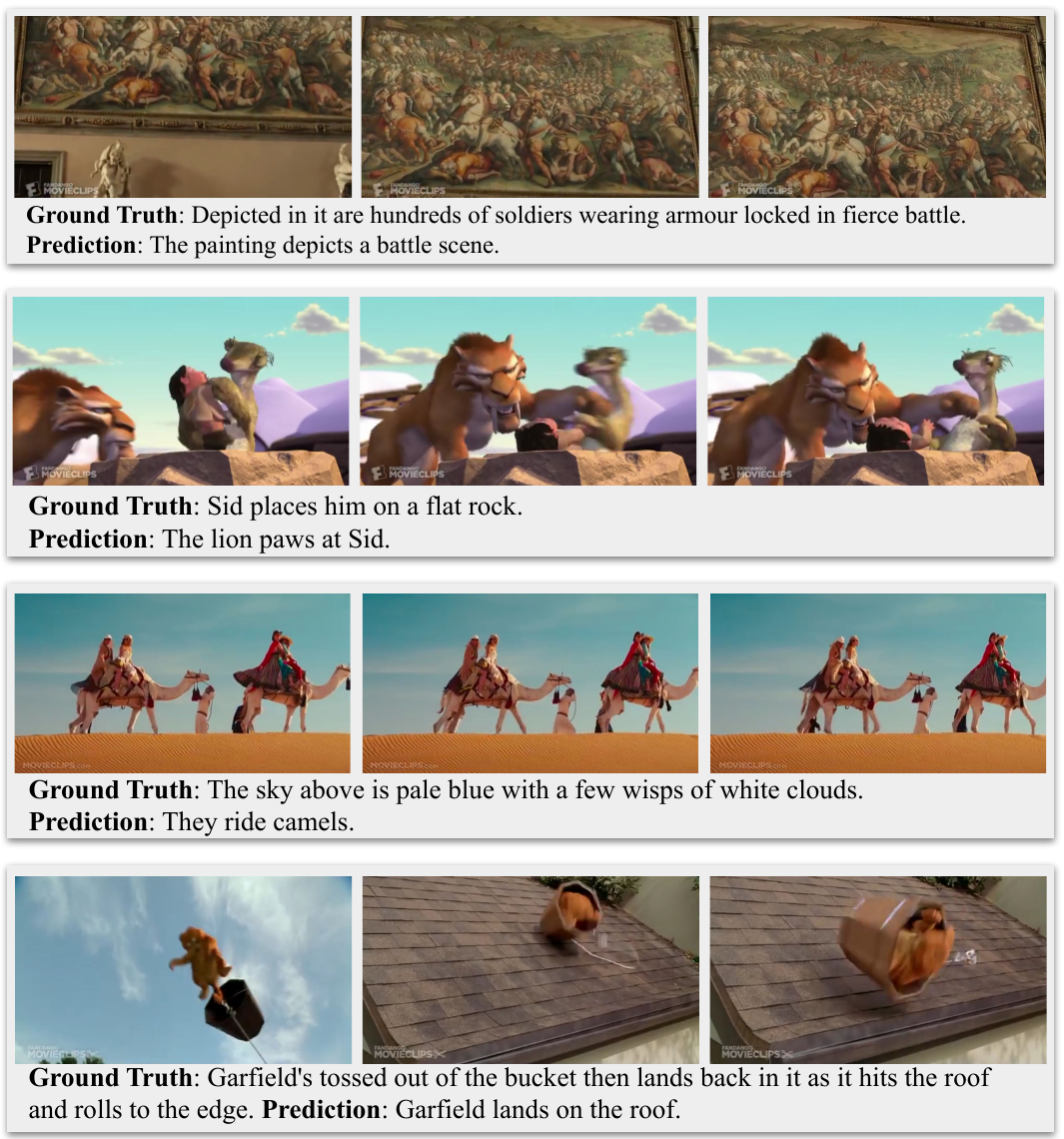}
    \caption{
    Additional qualitative results. For each movie clip, three frames are uniformly extracted and visualized. The movie clips are from Inferno (2016, \texttt{tt3062096}), 
    Ice Age (2002, \texttt{tt0268380}), 
    Sex and the City 2 (2010, \texttt{tt1261945}) 
    and Garfield (2004, \texttt{tt0356634}).
    For qualitative results in video format, please refer to the \texttt{mp4} file in our supplementary material.
    }
    \label{fig:supp:qual}
\end{figure}

\subsection{Analysis of the Evaluation Metrics}

A few examples of AD annotations and (both good and bad) model predictions are shown in Figure~\ref{fig:supp:metrics}, we show both CRITIC and LLM-AD-Eval metrics for each pair.
Qualitatively, we can feel that LLM-AD-Eval correlates to human judgement of semantic matching between two sentences.
The CRITIC metric gives zeros when the predictions do not contain the characters in the annotation.

\subsection{Better Character Exemplars for CMD-AD}

In AutoAD-II~\cite{Han23a}, the character exemplars are constructed from the in-context nearest neighbours of actors' IMDb profile pictures. However, we qualitatively find this method does not work well for old movies where the IMDb profile pictures are of low quality, or cartoon movies where the IMDb profile pictures are the voice actors instead of the characters.

We investigated an additional source of character exemplars, which are from the `photo' pages on IMDb, like~\url{https://www.imdb.com/title/tt2294629/characters/nm0068338}
or~\url{https://www.imdb.com/title/tt0046912/characters/nm0001537}.
These photos are the movie frames uploaded by users, corresponding to specific characters. However, this source has two major issues: first, the photos are biased towards famous characters, and less-known characters usually do not have photos uploaded; second, photos under one character may contain other characters.
We find only 50\% characters have valid photo pages, and we use face detector\footnote{\url{https://github.com/ageitgey/face_recognition}} to select photos with only one face;
For those without photo pages, we fall back to the original AutoAD-II strategy to find in-context exemplars.
The main paper uses this new strategy for character exemplars for all the experiments.
We ablate the effect of this update on the character exemplar in Table~\ref{tab:supp:char}.

\begin{table}[t]
    \centering
    \setlength{\tabcolsep}{3pt}
    \begin{tabular}{ll|cc}
    \toprule
        Method         & Setting            & CIDEr & CRITIC  \\ \midrule
        Movie-Llama2   & new exemplars      & \textbf{21.7}  & \blue{\textbf{25.2}}    \\ 
        Movie-Llama2   & old exemplars      & 20.1  & \blue{23.3}    \\ 
    \bottomrule
    \end{tabular}
    \vspace{-2mm}
    \caption{\textbf{Effect of new character exemplars} on CMD-AD-Eval.
    Both experiments are conducted without HowTo-AD pretraining.
    }
    \vspace{-2mm}
    \label{tab:supp:char}
\end{table}

\section{Additional Qualitative Results}

More qualitative results are shown in Figure~\ref{fig:supp:qual}.
Additionally, on the project page~\url{https://www.robots.ox.ac.uk/vgg/research/autoad/}, we display a few visualization of our model predictions as MP4 files. The AD ground-truth and our model predictions are provided as subtitles with the format `[title] ground-truth $\|$ prediction'.
The voice of AD is generated from OpenAI text-to-speech API\footnote{\url{https://platform.openai.com/docs/guides/text-to-speech}} and fused with the original movie soundtrack.

\input{fig/aa_match}
\input{fig/llm_eval}

%% file: fig/aa_match.tex
\newcommand\algcomment[1]{\def\@algcomment{\footnotesize#1}}
\begin{algorithm*}[h]
\caption{Python script for audio-audio alignment.}\label{code:aa_match}
\begin{lstlisting}[language=python]
import torch
import torchaudio
import numpy as np
from sklearn.linear_model import RANSACRegressor
from sklearn.metrics import mean_squared_error

def aa_match(cmd_wav, audiovault_wav, audiovault_anno):
    """
    cmd_wav: audio path of CMD 
    audiovault_wav: audio path of AudioVault chunk,
    audiovault_anno: AD annotations (with timestamps) of the AudioVault chunk
    
    Return: mse, slope and intercept of RANSAC
    """
    resolution_per_mel = 512/16000

    cmd_waveform, cmd_sr = torchaudio.load(cmd_wav)
    cmd_melspec = mel_spectrogram(cmd_waveform)
    cmd_melspec = cmd_melspec / cmd_melspec.norm(dim=-2, keepdim=True)
    
    av_waveform, av_sr = torchaudio.load(audiovault_wav)
    av_melspec = mel_spectrogram(av_waveform)
    av_melspec = av_melspec / av_melspec.norm(dim=-2, keepdim=True)

    # mask melspec as zeros if AD appears. Only compute non-AD part
    av_melspec_masked = torch.clone(av_melspec)
    start_end_mel_idx = []
    for _, row in audiovault_anno.iterrows():
        start_mel_idx = int(row['start'] / resolution_per_mel) # second to mel idx
        end_mel_idx = int(row['end'] / resolution_per_mel)
        av_melspec_masked[...,start_mel_idx:end_mel_idx] = 0
        start_end_mel_idx.append([start_mel_idx, end_mel_idx])
    
    window_size = 50  # equal to 3.2 second

    # use 1D conv to compute correlation in parallel
    # use unfold() to get chunks by moving windows
    melspec_av_all_cuts = av_melspec_masked.unfold(dimension=-1, size=window_size, step=window_size)
    melspec_unfold = cmd_melspec.unfold(dimension=-1, size=window_size, step=1)
    conv_output = torch.einsum('bctw,bcaw->bta', melspec_unfold, melspec_av_all_cuts)
    ad_mask = torch.any(melspec_av_all_cuts.mean(1)==0, dim=-1)
    conv_output = conv_output[0, :, ~ad_mask[0].bool()]  # only keep non-AD chunks
    max_cor, max_pos = conv_output.max(0)
    plot_outputs = torch.stack(
        [torch.arange(0, av_melspec_masked.shape[-1]-window_size+1, window_size)[~ad_mask[0].bool()],
         max_pos, max_cor/window_size], -1).numpy()

    # prepare scatters for RANSAC
    peak_cor = np.max([i[-1] for i in plot_outputs])
    samples_X = (plot_outputs[:, 1:2] + np.linspace(0, window_size, 5)[None,:]).flatten()
    samples_y = (plot_outputs[:, 0:1] + np.linspace(0, window_size, 5)[None,:]).flatten()
    samples_weight = plot_outputs[:, 2:3].repeat(5, axis=1).flatten()

    ransac = RANSACRegressor(random_state=0, residual_threshold=50).fit(
        samples_X[:,None], samples_y[:,None], samples_weight)

    inlier_mask = ransac.inlier_mask_
    outlier_mask = np.logical_not(inlier_mask)
    slope = ransac.estimator_.coef_[0,0]
    intercept = ransac.estimator_.intercept_[0]

    # get mse
    y_pred = ransac.predict(samples_X[inlier_mask][:,None])
    mse = mean_squared_error(samples_y[inlier_mask][:,None], y_pred)

    print('MSE=', mse)
    print('slope=', slope)
    print('intercept=', intercept)
    return mse, slope, intercept

\end{lstlisting}
\end{algorithm*}

%% file: fig/llm_eval.tex
\renewcommand\algcomment[1]{\def\@algcomment{\footnotesize#1}}
\begin{algorithm*}[h]
\caption{Python script for LLM-AD-Eval, with OpenAI gpt-3.5-turbo API.}\label{code:llm_ad_eval}
\begin{lstlisting}[language=python]
import ast
import openai

def eval_each(text_gt, text_pred, client):
    # Compute the LLM-AD-Eval score
    completion = client.chat.completions.create(
        model="gpt-3.5-turbo",
        messages=[
            {
                "role": "system",
                "content":
                    "You are an intelligent chatbot designed for evaluating the quality of generative outputs for movie audio descriptions. "
                    "Your task is to compare the predicted audio descriptions with the correct audio descriptions and determine its level of match, considering mainly the visual elements like actions, objects and interactions. Here's how you can accomplish the task:"
                    "------"
                    "##INSTRUCTIONS: "
                    "- Check if the predicted audio description covers the main visual events from the movie, especially focusing on the verbs and nouns.\n"
                    "- Evaluate whether the predicted audio description includes specific details rather than just generic points. It should provide comprehensive information that is tied to specific elements of the video.\n"
                    "- Consider synonyms or paraphrases as valid matches. Consider pronouns like 'he' or 'she' as valid matches with character names. Consider different character names as valid matches. \n"
                    "- Provide a single evaluation score that reflects the level of match of the prediction, considering the visual elements like actions, objects and interactions."
            },
            {
                "role": "user",
                "content":
                    "Please evaluate the following movie audio description pair:\n\n"
                    f"Correct Audio Description: {text_gt}\n"
                    f"Predicted Audio Description: {text_pred}\n\n"
                    "Provide your evaluation only as a matching score where the matching score is an integer value between 0 and 5, with 5 indicating the highest level of match. "
                    "Please generate the response in the form of a Python dictionary string with keys 'score', where its value is the matching score in INTEGER, not STRING."
                    "DO NOT PROVIDE ANY OTHER OUTPUT TEXT OR EXPLANATION. Only provide the Python dictionary string. "
                    "For example, your response should look like this: {'score': }."
            }
        ]
    )
    # Convert response to a Python dictionary.
    response_message = completion.choices[0].message.content
    response_dict = ast.literal_eval(response_message)
    return response_dict

client = OpenAI()
eval_each(text_gt, text_pred, client)

\end{lstlisting}
\end{algorithm*}